\documentclass[letterpaper, 10 pt, conference]{ieeeconf}  

\IEEEoverridecommandlockouts                              
\usepackage{amsmath}
\usepackage{amssymb}
\usepackage{graphicx}
\usepackage{booktabs}
\usepackage{multirow}
\usepackage{bm}
\usepackage{caption}
\usepackage{subcaption}
\usepackage{tabularx}
\usepackage{wrapfig}
\usepackage{xcolor}

\newif\ifshowrevised
\showrevisedfalse 

\ifshowrevised
  \newcommand{\revised}[1]{\textcolor{magenta}{#1}}
\else
  \newcommand{\revised}[1]{#1}
\fi
\usepackage{makecell}

\title{\LARGE \bf
DOT-Sim: Differentiable Optical Tactile Simulation with Precise Real-to-Sim Physical Calibration
}

\author{
Yang You$^{1}$,
Won Kyung Do$^{1}$,
Aiden Swann$^{1}$,
Rika Antonova$^{1,2}$,
Monroe Kennedy$^{1}$,
Leonidas Guibas$^{1}$%
\thanks{$^{1}$Stanford University, United States. Email: 
{\tt\small \{yangyou, wkdo, swann, monroek, guibas\}@stanford.edu}}%
\thanks{$^{2}$University of Cambridge, United Kingdom.  Part of this work was completed while the author was with Stanford University. Email: 
{\tt\small rika.antonova@cst.cam.ac.uk}%
}
\thanks{Leonidas Guibas and Yang You acknowledge support from the Toyota Research Institute University 2.0 Program, ARL grant W911NF-21-2-0104, a Vannevar Bush Faculty Fellowship, and a gift from the Flexiv corporation. Yang You is also supported in part by the Outstanding Doctoral Graduates Development Scholarship of Shanghai Jiao Tong University. Aiden Swann is supported by NSF GRFP Fellowship No. DGE-2146755. This work was also supported in part by NSF Grant No. 2142773 and 2220867.}
}

\begin{document}

\maketitle
\thispagestyle{empty}
\pagestyle{empty}
\begin{abstract}
Simulating optical tactile sensors presents significant challenges due to their high deformability and intricate optical properties. To address these issues and enable a physically accurate simulation, we propose \mbox{\textbf{DOT-Sim}}: \textit{\textbf{D}ifferentiable \textbf{O}ptical \textbf{T}actile \textbf{Sim}ulation}. Unlike prior simulators that rely on simplified models of deformable sensors, DOT-Sim accurately captures the physical behavior of soft sensors by modeling them as elastic materials using the Material Point Method (MPM).
DOT-Sim enables rapid calibration of optical tactile sensor simulation using a small number of demonstrations within minutes, which is substantially faster than existing methods. Compared to current baselines, our approach supports much larger and non-linear deformations.
To handle the optical aspect, we propose a novel approach to simulating optical responses by learning a residual image relative to the real-world idle state. We validate the physical and visual realism of our method through a series of zero-shot sim-to-real tasks.
Our experiments show that \mbox{DOT-Sim} (1) accurately replicates the physical dynamics of a DenseTact optical tactile sensor in reality, (2) generates realistic optical outputs in contact-rich scenarios, and (3) enables direct deployment of simulation-trained classifiers in the real world, achieving 85\% classification accuracy on challenging objects and 90\% accuracy in embedded tumor-type detection, and (4) allows precise trajectory following with policy trained from demonstrations in simulation with an average error of less than 0.9 mm.
\end{abstract}


\begin{figure*}[ht]
    \centering
    \vspace{0.5em}
    \includegraphics[width=0.85\linewidth]{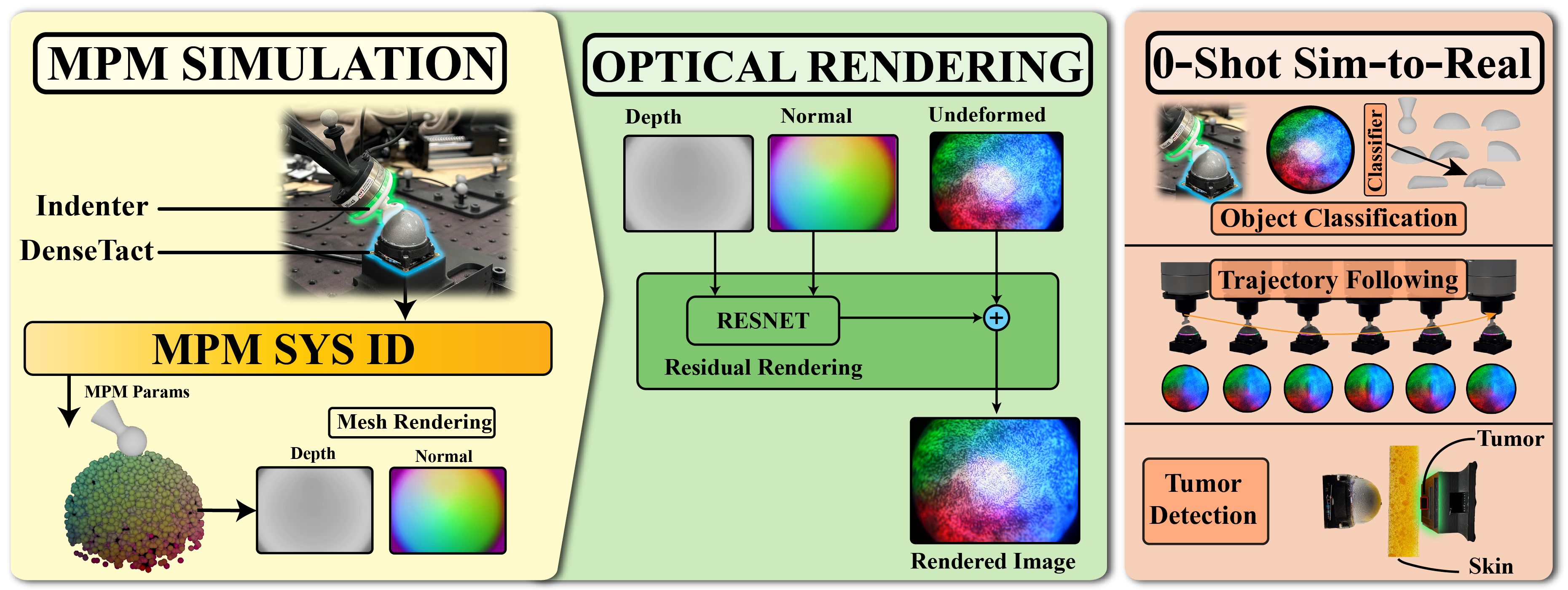}
    \caption{Optical tactile sensors with flexible surface materials are challenging to simulate due to complex physical deformation and internal optical effects. We propose DOT-Sim -- a framework that enables physically grounded and realistic tactile image generation by modeling both deformation and optical appearance. This enables zero-shot sim-to-real transfer for tasks such as indenter classification, trajectory following, and tumor detection.}
    \label{fig:overview}
    \vspace{-2em}
\end{figure*}

\section{Introduction}

Tactile sensing plays a critical role in enabling robots to perceive and interact with their environment in a physically grounded manner. Among various tactile modalities, optical tactile sensors such as GelSight~\cite{yuan2017gelsight} and DenseTact~\cite{do2022densetact,do2023densetact} have gained increasing popularity due to their high spatial resolution and ease of fabrication. However, the soft and deformable nature of these sensors, combined with their complex internal light transport mechanisms, poses significant challenges for simulation~\cite{si2024difftactile}. Accurate simulation of tactile sensors is essential for data generation, algorithm training, and sim-to-real transfer, yet remains underexplored.

Existing tactile simulation frameworks~\cite{do2022densetact,wang2022tacto,si2021taxim,agarwal2021simulation} often rely on simplified approximations of contact geometry or assume fixed mappings from force to image output. This is problematic, since many optical tactile sensors with flexible surface materials exhibit highly nonlinear responses to contact, due to both material elasticity and internal optical effects. Capturing these aspects with high fidelity is critical for enabling robust downstream applications, including perception, classification, and control. Hence, existing methods fall short in this regard.

To overcome these challenges, we introduce \textbf{DOT-Sim}:  \textit{\textbf{D}ifferentiable \textbf{O}ptical \textbf{T}actile \textbf{Sim}ulation} -- a physically grounded, differentiable simulation framework tailored for optical tactile sensors. Our method proceeds in two stages. First, we model the physical behavior of a tactile sensor using the Material Point Method (MPM), a particle-based continuum simulation approach well-suited for soft, elastic materials. We calibrate the physical parameters of the sensor, such as Young’s modulus and Poisson’s ratio, by aligning simulated deformations with a small number of real-world tactile observations. This calibration is made efficient through the use of differentiable simulation and gradient-based optimization.
Second, we model the sensor’s optical response by simulating its depth and surface normals via virtual camera rendering, 
and then learn a residual mapping from the simulated rendering to the real tactile image. Rather than predicting the raw image, we predict only the difference from an idle frame, thereby significantly improving the efficiency of learning to generate realistic tactile images.

Through extensive experiments, we show that DOT-Sim: 1) replicates the physical behavior of a recently developed DenseTact sensor with high fidelity; 2) generates realistic optical renderings in a variety of contact scenarios; 3) enables successful zero-shot sim-to-real transfer for downstream tasks, such as classification and trajectory following.
%
DOT-Sim improves over the strongest baseline by 17.34\% in average PSNR for optical image quality, achieves absolute accuracy gains of 28.24\% \& 44.83\% on unseen indenter \& tumor-type classification tasks respectively, and enables precise trajectory following in reality by controllers trained from demonstrations in simulation,
with an average error of $<0.9\,\mathrm{mm}$.
Our framework opens new possibilities for employing tactile sensing, especially in scenarios that require extra precision for perception and interaction.


\section{Background and Related Work}

\paragraph{Optical Tactile Sensors}
Optical tactile sensors such as GelSight~\cite{yuan2017gelsight,johnson2009retrographic}, DIGIT~\cite{lambeta2020digit}, and DenseTact~\cite{do2022densetact,do2023densetact} have become popular due to their high spatial resolution and rich contact feedback. They have been used in a variety of downstream tasks including object recognition~\cite{gao2016deep}, manipulation~\cite{qi2023general}, and shape reconstruction~\cite{yuan2017gelsight, swann2024touchgs}.
Such sensors encode surface deformation as images, enabling compatibility with standard vision-based learning techniques.
%
%
%
Despite variations in physical form factors, gel elasticity, and calibration methods, optical tactile sensors share a common structure: a soft deformable gel sensing element, an internal light emitting module, and camera with appropriate lenses (to observe the deformation of the gel sensing element). The gel element is built from clear gel materials (e.g. PDMS, silicone, or thermal plastics) coated with a reflective material to simplify sensing and capture the sensor deformations robustly. 
Markers or randomized patterns may be applied between the clear gel materials and the reflective surface. The camera captures sensor surface deformation, and the resulting images are used to extract shape information and contact force distributions. 
The complex interplay between deformation, lighting, and gel  properties enables these images to be highly expressive, but also creates modeling and simulation challenges.


\begin{table}[ht!]
    \centering
    \vspace{0.7em}
    \resizebox{\linewidth}{!}{
    \begin{tabular}{@{} l l l l l @{}}
        \toprule
        \textbf{Simulator} &
        \textbf{Phy. Sim} &
        \textbf{Backend} &
        \textbf{Optical Sim} &
        \textbf{Sim$\!\rightarrow\!$Real} \\
        \midrule
        Tacto \cite{wang2022tacto}                                   & PyBullet        & PyBullet  & OpenGL   & None \\[0.2em]
        Taxim \cite{si2021taxim}                              & FEM             & N/A  & Calibrated LUT & None \\[0.2em]
        DiffTactile \cite{si2024difftactile}       & FEM             & Taichi   & Learned reflectance    & Marker only \\[0.2em]
        DOT-Sim (ours)     & MPM             & Warp   & ResNet-based model      & Full optical \\ 
        \bottomrule
    \end{tabular}
    }
    \caption{Overview of the key existing optical tactile simulators and comparison with our work. We use a differentiable physical model and a fully learned optical model to achieve efficient and accurate calibration (real-to-sim alignment). Note \textbf{Sim$\!\rightarrow\!$Real} refers to policy learning, not only image generation. This table includes the most notable methods. }
    \vspace{-1em}
    \label{tab:tactile_sim_simple}
\end{table}

\paragraph{Physical Modeling}
Tacto~\cite{wang2022tacto} is a widely used rendering-based tactile simulator that approximates contact using analytical geometry and heuristics for visual appearance. While computationally efficient, it lacks physical realism and fails to generalize to complex contacts or deformations. In~\cite{xu2022efficient}, a penalty-based collision model is used to improve simulation speed at the cost of lower accuracy. Diff-Tactile \cite{si2024difftactile} uses a finite element method (FEM) to preform physics simulation of the deformable sensor. DiffTaichi~\cite{hu2019difftaichi} and ChainQueen~\cite{hu2019chainqueen} introduced differentiable simulation frameworks using the Material Point Method (MPM), but were primarily targeted at general-purpose deformable body simulation and not optimized for tactile sensors. Our approach builds upon MPM but focuses on dense, real-world calibration of tactile dynamics and optical appearance.


\paragraph{Optical Simulation}
Rendering the optical properties of a sensor is very difficult due to the complex reflective properties and shadows which occur within the sensor. Many works choose to forgo optical simulation entirely and instead simulate simplified proxies, such as markers on the sensor surface, when obtaining policies for sim-to-real tasks~\cite{xu2022efficient, si2024difftactile, zhao2024fotsfastopticaltactile}. Several recent works utilize simplified optical models to render the sensor. In Tacto~\cite{wang2022tacto}, sensor images are rendered using OpenGL. These images can then be used to augment the real image. Taxim~\cite{si2021taxim} utilizes a polynomial lookup table to color its sensor images. While these methods are fast, they lack visual accuracy. 
Several simulators use learning for optical simulation \cite{si2024difftactile, zhao2024fotsfastopticaltactile} by either learning to generate images from scratch, or learning one component of an otherwise analytical rendering method. While these methods do employ learning for image generation, the resulting images are not accurate enough for policy conditioning. Instead, they utilize marker motion, which is easier to simulate and can be tracked on the real sensor.
This compromises the expressive signal produced by the sensor and ultimately reduces the set of tasks which can be trained in sim. Furthermore, these methods learn small MLP neural networks, which operate on a per-pixel basis. In contrast, our method uses a single ResNet, which outputs the entire rendered image.
High accuracy can be achieved by applying GANs to the output of a lower fidelity simulator~\cite{chen2022bidirectional}. However, the indenter set used in~\cite{chen2022bidirectional} is designed to be very easy to discriminate. Furthermore, \cite{chen2022bidirectional} solves a fundamentally easier problem since the visual output of the GelSight used there is uniform, compared to the DenseTact's pattern surface. Moreover, our method provides a full pipeline including physical and optical simulations.

%



\paragraph{System Identification and Sim-to-Real Transfer for Tactile Perception}
Several recent works have explored learning tactile dynamics in a differentiable manner to facilitate calibration (system identification), e.g. PhysDreamer~\cite{zhang2024physdreamer} and SpringGaus \cite{zhong2024springgaus}. DiffTactile~\cite{si2024difftactile} applies this method to tactile sensing, utilizing a differentiable FEM simulation. DiffTactile optimizes their FEM simulation based only on tactile marker pose tracking and force input across a few trajectories. In contrast, we rely on a high fidelity Abaqus model, which generates full deformed meshes across several thousand indentations, enabling more direct supervision. 

Table \ref{tab:tactile_sim_simple} gives a summary of recent tactile simulators and contrasts them with our proposed method. 

\section{The Proposed Approach: DOT-Sim}

Simulation of optical tactile sensors is challenging due to the manufacturing variability of  physical components (e.g. the gel elements) and the complexity of accurately modeling optical components. \linebreak
%
To achieve accurate simulation, we propose a two-stage framework that integrates physically grounded modeling with data-driven optical rendering, as illustrated in Figure~\ref{fig:overview}. Instead of directly learning a mapping from forces to tactile images, we embed strong physical priors by first modeling the sensor’s mechanical deformation using the Material Point Method (MPM). We then calibrate the sensor’s physical parameters via differentiable physics, leveraging a small number of real-world interactions between the sensor and objects. This process yields paired data of sensor images and corresponding deformed sensor meshes, which can be obtained from the standard software provided with most optical tactile sensors.

After obtaining a physically accurate model, we simulate the sensor's mechanical deformation under forces from interaction with objects. To replicate optical rendering, we position a virtual camera inside the simulated sensor to render depth and surface normals. Then, we input these into a neural network that predicts a residual optical image relative to a reference (idle) frame. This substantially reduces approximation error compared with direct image regression.


\subsection{Calibration of the Physical Properties of Tactile Sensors}
\label{sec:physics}

To simulate the mechanical behavior of a tactile sensor, we model it as a collection of particles governed by the Material Point Method (MPM); \cite{jiang2016material} provides a recent overview of~MPM. Each particle represents a small volume of the sensor and carries physical properties including volume $V_p$, mass $m_p$, position $\bm{x}_p^t$, velocity $\bm{v}_p^t$, deformation gradient $\bm{F}_p^t$, and local velocity field gradient $\bm{C}_p^t$ at time step $t$. MPM is a hybrid Eulerian-Lagrangian method, which also maintains another set of grid node mass $m_i$ and velocity $\bm{v}_i$, where $i$ is the grid index.
We consider the case where the tactile sensor is in contact with rigid indenters. Let $\tilde{\bm{x}}^t$ and $\tilde{\bm{v}}^t$ denote the position and velocity of the indenter at time $t$. The dynamics of the full system are:
\begingroup
\small
\begin{align}
    \{m_i\},\{\bm{v}_i\} := P2G\big(\{m_p\}, \{\bm{x}_p^t\}, \{\bm{v}_p^t\}, \{\bm{F}_p^t\}, \{\bm{C}_p^t\}, \theta, \Delta t\big) \\
    \{m_i\}, \{\bm{v}_i\} := CM\big(\{m_i\}, \{\bm{v}_i\}, \tilde{\bm{x}}^t, \tilde{\bm{v}}^t, \theta, \beta, \Delta t\big) \\
    \{\bm{x}_p^{t+1}\}, \{\bm{v}_p^{t+1}\}, \{\bm{F}_p^{t+1}\}, \{\bm{C}_p^{t+1}\} := G2P\big(\{m_i\}, \{\bm{v}_i\}, \theta, \Delta t\big) \\
    \tilde{\bm{x}}^{t+1}, \tilde{\bm{v}}^{t+1} := FK(\tilde{\bm{x}}^t, \tilde{\bm{v}}^t, \beta, \Delta t).
\end{align}
\endgroup

Here, $P2G$ and $G2P$ denote the MPM \textit{particle to grid} and \textit{grid to particle} transfer operations. $CM$ is a contact model of the sensor-indenter interaction dynamics, $FK$ is the forward kinematics function for the indenter.
Parameter vector $\theta \!=\! (E,\nu)$ encapsulates the physical properties of the sensor. In this work, we focus on calibrating Young’s modulus $E$ and Poisson’s ratio $\nu$, following prior work such as PhysDreamer~\cite{zhang2024physdreamer}. Indenter properties $\beta$ are fixed, thus treating the indenter as rigid and unaffected by contact forces to reflect real-world conditions (where indenters are typically actuated by robot arms or human hands, and are not influenced by sensor deformation). 

To estimate $\theta\!=\!(E, \nu)$, we collect a small set of real-world demonstration sequences (19 videos), recording the indenter position \revised{in a calibrated OptiTrack marker tracking system}, and then generate the pseudo–ground-truth mesh deformation with the Abaqus 2024 Finite Element Analysis (FEA) simulator \cite{Dassault2024Abaqus}, which provides accurate but computationally expensive deformation results.
\revised{We rely on Abaqus to generate pseudo ground-truth because directly capturing the deformed point cloud during indentation is highly challenging due to severe occlusions from the indenter, limited depth-sensor resolution, and measurement noise. Following prior works \cite{zhao2024ifem2,do2025tensortouch}, we align our MPM simulator with Abaqus outputs, which have been demonstrated to be accurate. For Abaqus simulations, we specify physical parameters such as the silicone–plastic friction coefficient and Yeoh hyperelastic material constants (for the gel model). These are obtained through standardized uniaxial and biaxial tests based on Densetact specifications \cite{ogden1997non}. Notably, these parameters are distinct from the learnable $\theta$ in our MPM model. Since the Abaqus solver is slow, complex, and lacks GPU acceleration, it is impractical to use it directly as the physical simulator, especially in reinforcement learning where simulation data must be generated online. Instead, we calibrate our MPM simulator to match Abaqus’ accurate but costly deformations, enabling efficient and scalable physics simulation.}

\begin{figure}
    \centering
    \vspace{0.5em}
    \includegraphics[width=0.7\linewidth]{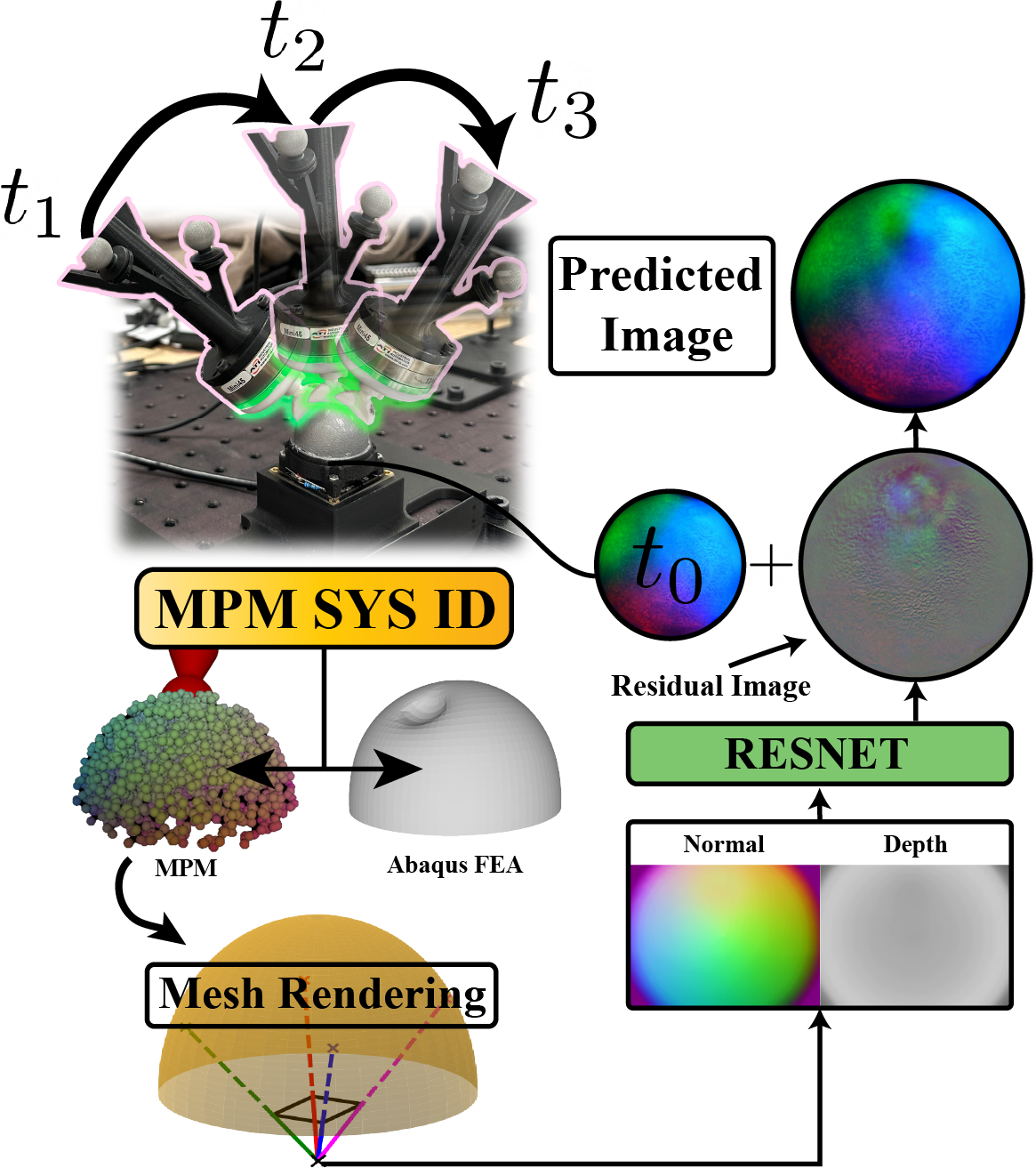}
    \caption{Sensor calibration.}
    \vspace{-2em}
    \label{fig:calibration}
\end{figure}

 We then optimize $\theta$ by differentiable minimization of the Chamfer distance between the simulated and real point clouds. The trajectories are collected by poking the sensor at various angles and depths, and differentiable MPM simulations are run in parallel for all sequences. To obtain a robust estimate, we take the median of the estimated $(E, \nu)$ values over all sequences.
 The calibration process is highly efficient, completing within a few minutes on a single A5000 GPU, significantly faster than previous approaches~\cite{do2023densetact}. Figure~\ref{fig:calibration} illustrates the calibration process. For the differentiable minimization, we re-parameterize $E$ by optimizing $\log(E)$ instead. Both $\log(E),\nu$ have a learning rate of 0.1, and are optimized for 30 iterations.


\subsection{Optical Simulation: A Residual Approach}
\label{sec:optical}
While MPM simulation provides accurate modeling of physical deformation, the final output of a real-world tactile sensor is an optical image. Directly simulating light transport through the elastomer is challenging due to complex and sensor-specific optical properties, such as non-uniform gel coloration and internal reflections.
To address this, we propose a hybrid approach that combines physics simulation for geometry with a neural rendering model for optics. Specifically, we first render depth and normal maps from the simulated sensor by placing a virtual camera at the center of the sensor base and casting rays through the upper hemisphere, mimicking the imaging process of real-world tactile sensors (see bottom left part of Figure~\ref{fig:overview}). Given these simulated depth and normal maps, we input them into a neural network (with the U-Net architecture) that predicts the corresponding optical RGB image (middle part of Figure~\ref{fig:overview}).

Although directly predicting the RGB image is feasible, it does not exploit the key observation that most regions of the tactile image remain static during contact, with deformation-induced signals manifesting primarily as localized residuals. Therefore, we predict a \emph{residual image} relative to an idle (contact-free) reference frame. This residual image is defined as the difference between a contact frame and the idle frame, i.e., the frame at $t\!=\!0$ when the indenter is not in contact with the sensor. We add the residual image to the real-world idle image to obtain the final optical output.


We employ the PyTorch implementation of \textit{DeepLabV3 ResNet50}~\cite{chen2017rethinking} as our backbone. The training objective is a simple $\ell_2$ loss computed on pixel-wise RGB values. For all experiments, we use a batch size of 8 and a learning rate of 3e-4, with a weight decay of 1e-4, using the Adam optimizer. The model is trained for 100 epochs.


\section{Experimental Results} \label{sec:results}

We evaluate DOT-Sim through a series of experiments designed to assess the realism and effectiveness of both its physical simulation and optical rendering.
Section~\ref{sec:physical_accuracy} evaluates the accuracy of the physical simulation.
Section~\ref{sec:optical_accuracy} assesses the fidelity of the optical simulation by comparing rendered outputs with real-world tactile images, and also presents an ablation comparing  to direct regression without residual prediction.
Section~\ref{sec:classification_tasks} shows performance in downstream tasks, such as indenter classification, tumor detection, and trajectory tracking with sim-to-real transfer.

Since \mbox{DOT-Sim} can handle large sensor deformation, for our experiments we selected the \mbox{DenseTact 2.0} sensor, which is a soft tactile sensor that can undergo significant deformation. Smaller deformations would be easier to handle, so the method should also work well with sensors that experience deformations smaller than those of the DenseTact 2.0. Furthermore, DenseTact 2.0 has a randomized pattern on its surface, which makes it particularly challenging to simulate with existing optical or physical simulators.
Like most optical tactile sensors, DenseTact 2.0 consists of a camera, camera-gel mount, and the gel to sense deformation. The hemispherical gel is made of clear silicone, coated with a randomized pattern and reflective surface (this  enables effective observation of contact-induced deformations). Given an optical image of the deformation, the corresponding deformed sensor mesh is generated by finite element analysis of the gel. This provides near ground-truth deformation of the sensor given a known indenter pose and shape.

\subsection{Physical Accuracy of 3D Deformation} \label{sec:physical_accuracy}

We evaluate the physical realism of the simulated sensor deformation by comparing the particle-based simulated surface against the observed DenseTact point cloud across various indenters. This evaluation is conducted using the evaluation dataset from all indenters in Medium (M) setting.

\paragraph{Evaluation Metrics} Following the evaluation protocols in~\cite{yuan2018pcn, xie2020grnet}, we report performance using four standard metrics: L2 Chamfer Distance (L2 CD), Significant L2 Chamfer Distance (Sig. L2 CD), Earth Mover’s Distance (EMD), and F-Score at 1mm. For each evaluation, we uniformly sample 2,048 points from both the predicted and ground-truth point clouds.
L2 Chamfer Distance measures the average normed distance between each point in one point cloud and its nearest neighbor in the other, serving as an overall geometric similarity. Significant L2 Chamfer Distance is a variant of L2 considering only the top 1\% furthest nearest neighbor for each point. Earth Mover's Distance computes the minimum cost of transforming one point cloud into the other. F-Score at 1mm evaluates the harmonic mean of precision and recall under a 1mm threshold, indicating the accuracy of point-level correspondences. For baselines, we compare with DiffTactile,  Tacto and Taxim.

\begin{table}
\vspace{0.7em}
\centering
\begin{tabular}{lcccc}
\toprule
\textbf{Method} & \textbf{L2 CD} & \textbf{Sig. L2 CD}  & \textbf{EMD}  & \textbf{F-Score} \\
 & mm $\downarrow$ & mm $\downarrow$ &  mm $\downarrow$ &  @1mm $\uparrow$ \\
\midrule
*DiffTactile & - & - & - & - \\
Tacto & 1.77 & 4.21 & 1.33 & 63.59 \\
Taxim & 1.74 & 3.97 & 1.31 & 64.69\\
\textbf{DOT-Sim} & \textbf{1.71} & \textbf{3.82} & \textbf{1.29} & \textbf{69.89} \\
\bottomrule
\end{tabular}
\caption{Discrepancy between the simulated surface and the observed point cloud. Lower is better for CD and EMD; higher is better for F-Score. *DiffTactile: see explanation in the Results paragraph.}
\vspace{-2em}
\label{tab:realism_comparison}
\end{table}


\paragraph{Results}
Table~\ref{tab:realism_comparison} compares discrepancy between the simulated surface and the observed DenseTact point cloud. \mbox{DOT-Sim} achieves consistently better performance than both Taxim and Tacto across all metrics, demonstrating improved fidelity in modeling 3D surface deformation. \linebreak
Notably, we achieve a lower EMD (1.29 vs. 1.31 mm) and higher \mbox{F-Score at 1mm} (69.89 vs. 64.69), indicating a closer alignment. \revised{Note that L2 CD and EMD appear close to the baselines because they are averaged over the space, which masks the substantial improvement in the small regions directly under the indenter. Significant L2 error and F-Score compute error only over the largely deformed voxels, so are the more relevant metrics, and on these DOT-Sim shows a substantial reduction in error.} We attempted to use the open-source code released by the authors of DiffTactile, but observed that the FEM sensor simulation was overly stiff, showing no noticeable deformation in the provided examples. Most apparent `deformation' stemmed from the penalty-based contact model rather than true FEM response. Although the paper proposes a differential system identification method to tune FEM parameters, no implementation was included in the codebase, and the publication lacked sufficient detail for reproduction.

\subsection{Optical Simulation Accuracy}
\label{sec:optical_accuracy}  

\begin{figure}[ht]
    \centering
    \includegraphics[width=0.95\linewidth]{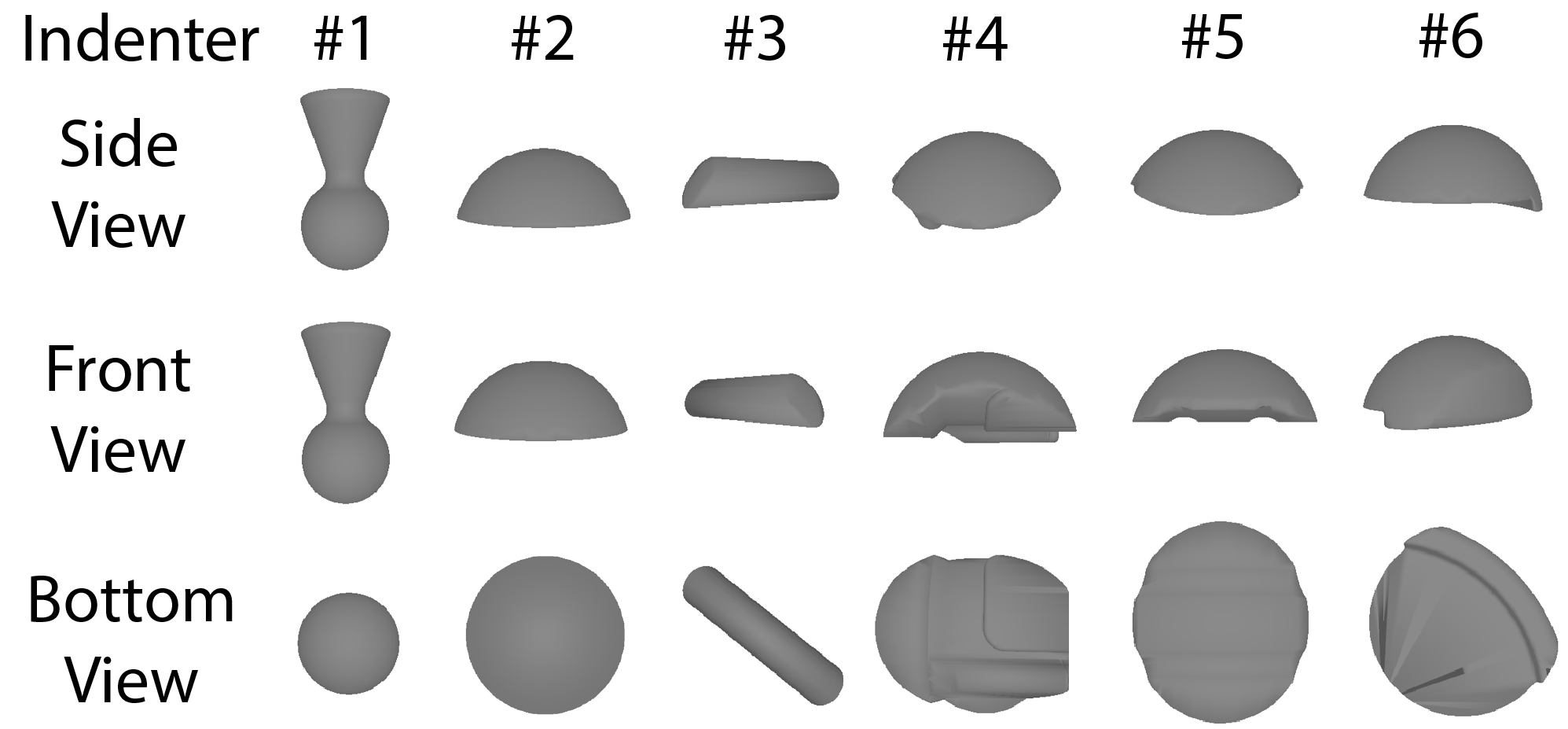}
    \caption{Indenters.}
    \label{fig:indenters}
\end{figure}

To quantify the accuracy of DOT-Sim’s optical outputs, we compare simulated and real tactile images in various scenarios. We use videos from six indenters captured at 24 FPS. Figure~\ref{fig:indenters} shows the indenters. We report the following metrics: \textbf{Mean L2 Norm} ($\downarrow$), \textbf{Significant Pixel L2 Norm} ($\downarrow$), and \textbf{Peak Signal-to-Noise Ratio (PSNR)} ($\uparrow$), where arrow direction shows whether higher or lower values are better (Appendix gives further details).

We evaluate on three levels of difficulty settings: \\
\noindent \textbf{Easy (E):} Two indenters (\#1, \#3) are used in both training and evaluation, with a random 80/20 split. 
\textbf{Medium (M):} All six indenters (\#1 -- \#6) are used with a random 80/20 training/evaluation split. \linebreak
\textbf{Hard (H):} A leave-two-out setup where four indenters (\#2, \#4, \#5, \#6) are used for training, and two unseen indenters (\#1, \#3) are used for evaluation.

\begin{table}[ht]
    \centering
    \vspace{0.7em}
    \begin{tabular}{@{}llccc@{}}
        \toprule
        \textbf{Setting} & \textbf{Method} & \textbf{Mean L2} $\downarrow$ & \textbf{PSNR} $\uparrow$ & \textbf{Sig. L2} $\downarrow$ \\
        & & $\times 10^{-2}$ & & $\times 10^{-2}$ \\
        \midrule
        \multirow{3}{*}{Easy} 
            & DiffTactile & 3.80 & 28.73 & 4.91 \\
            & Tacto (calib) & 4.92 & 26.97 & 6.18 \\
            & \textbf{DOT-Sim} & \textbf{2.85} & \textbf{32.12} & \textbf{3.68} \\
        \midrule
        \multirow{3}{*}{Medium} 
            & DiffTactile & 7.28 & 22.89 & 8.99 \\
            & Tacto (calib) & 5.35 & 26.35 & 6.71 \\
            & \textbf{DOT-Sim} & \textbf{3.25} & \textbf{31.39} & \textbf{4.21} \\
        \midrule
        \multirow{3}{*}{Hard} 
            & DiffTactile & 8.63 & 21.31 & 10.67 \\
            & Tacto (calib) & 5.04 & 26.79 & 6.35 \\
            & \textbf{DOT-Sim} & \textbf{3.50} & \textbf{30.48} & \textbf{4.53} \\
        \bottomrule
    \end{tabular}
    \caption{Comparison of DOT-Sim against baselines.}
    \vspace{-1.5em}
    \label{tab:optical_results}
\end{table}

\begin{figure}[ht]
    \centering
    \includegraphics[width=\linewidth]{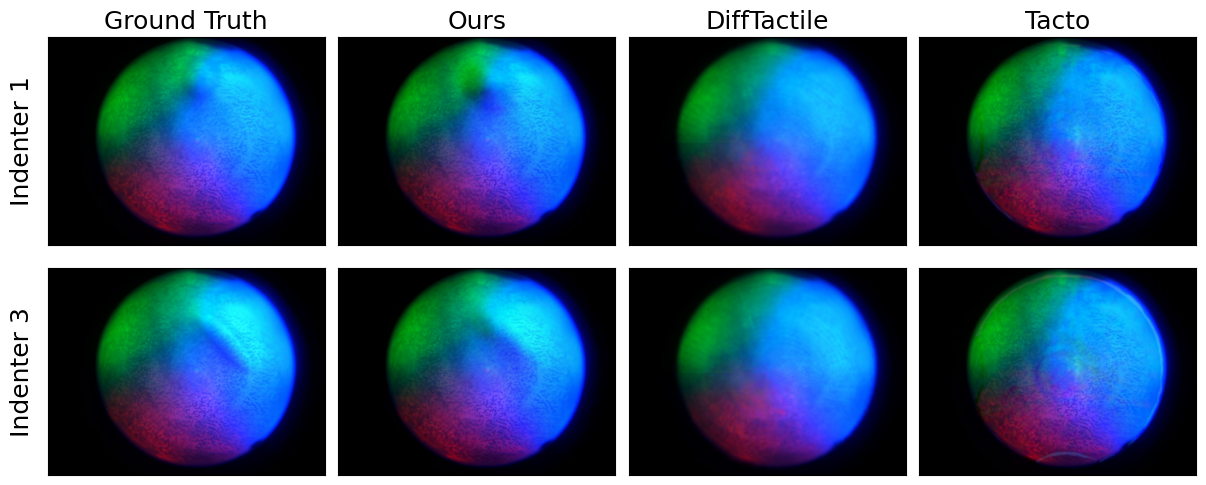}
    \caption{DOT-Sim captures fine-grained and anisotropic features more faithfully than prior methods in Hard (H) setting.}
    \vspace{-2em}
    \label{fig:comp}
\end{figure}

\paragraph{Results} We compare our method with DiffTactile~\cite{si2024difftactile} and Tacto~\cite{wang2022tacto}, Table \ref{tab:optical_results} shows the quantitative results. We observe significant improvements in visual accuracy: as high as 4 points of PSNR. For comparison, DiffTactile~\cite{si2024difftactile} reports 1 point PSNR improvement over their own baseline Taxim~\cite{si2021taxim}.  Figure~\ref{fig:comp} shows side-by-side comparisons of simulated and ground-truth tactile images under the Hard (H) setting. DOT-Sim improves over the strongest baseline by \textbf{17.34}\% on average PSNR. DOT-Sim faithfully captures fine-grained contact features, including anisotropic responses  (underrepresented in prior methods) that arise with indenter \#3, which exhibits anisotropic geometry.

\begin{figure}[hb]
    \centering
    \vspace{-2em}
    \includegraphics[width=0.85\linewidth]{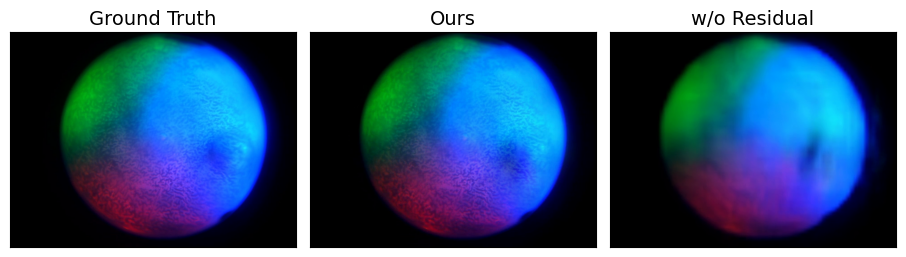}
    \caption{Qualitative comparison of our residual image prediction vs direct regression without residual learning.}
    \vspace{-1em}
    \label{fig:ablation}
\end{figure}

To evaluate the effectiveness of the proposed residual image prediction, we conduct an ablation study comparing DOT-Sim against a baseline that directly regresses the contact optical frame from the input normals and depths, without residual refinement. The evaluation is done under the Hard (H) setting.
Quantitative results are summarized in Table~\ref{tab:ablation}, and qualitative comparisons are shown in Figure~\ref{fig:ablation}. DOT-Sim achieves a higher PSNR (30.48 vs. 28.89) and lower significant L2 error ($4.53 \times 10^{-2}$ vs. $5.39 \times 10^{-2}$), confirming that the residual prediction significantly improves image fidelity, while direct regression produces blurry images.

\begin{table}[ht]
    \centering
    \begin{tabular}{lcc}
        \toprule
        \textbf{Method} & \textbf{PSNR} $\uparrow$ & \textbf{Sig. L2} $\downarrow$ \\
        & & $\times 10^{-2}$ \\
        \midrule
        w/o Residual & 28.89 & 5.39 \\
        \textbf{DOT-Sim} & \textbf{30.48} & \textbf{4.53} \\
        \bottomrule
    \end{tabular}
    \caption{Ablating residual image prediction.}
    \vspace{-2em}
    \label{tab:ablation}
\end{table}

\subsection{Performance in Downstream Classification, Detection, and Sim-To-Real Applications}
\label{sec:classification_tasks}

\paragraph{Sim-To-Real Zero-Shot Indenter Classification} To assess whether the simulated optical images preserve sufficient discriminative information, we design a classification task where a model is trained to identify the type of indenter solely from simulated tactile images. Importantly, evaluation is performed entirely on real-world tactile images, making this a stringent test of sim-to-real generalization.
We consider two settings:  a) \textit{in-domain} -- the classifier is trained on simulated images of indenters that were also used during residual image prediction training, i.e. indenters \#1 and \#3; \linebreak
b) \textit{out-of-domain} -- the classifier is trained on simulated images of indenters that were not used during residual image prediction training. 
For b), only the indenter meshes are available at test time, real images of these indenters are not seen during training; indenters \#2, \#4, \#5, and \#6 are used for residual image prediction training, while the classifier is trained on simulated images from indenters \#1 and \#3. In both settings, the classifier's training data is generated entirely from simulated optical outputs of our residual prediction module described in Section~\ref{sec:physics} and \ref{sec:optical}. This requires the simulated images to be not only realistic but also sufficiently aligned with the real-world domain for successful downstream task transfer.

\begin{table}
\centering
\vspace{0.7em}
\begin{tabular}{lcc}
\toprule
\textbf{Method} & \textbf{In-domain} & \textbf{Out-of-domain} \\
& Accuracy (\%) & Accuracy (\%) \\
\midrule
DiffTactile & 65.88 & 52.94 \\
*Tacto & 50.59 & 50.59 \\
\textbf{DOT-Sim} & \textbf{90.48} & \textbf{81.18} \\
\bottomrule
\end{tabular}
\caption{Indenter classification accuracy (\%) on real-world images using simulated data. *Tacto does not learn optical rendering, so performs the same across both settings (since only the training set differs between in- and out-of-domain).}
\label{tab:classification_results}
\vspace{-1em}
\end{table}

Results in Table~\ref{tab:classification_results} show that \mbox{DOT-Sim} significantly outperforms prior methods in both settings. Notably, in the challenging out-of-domain scenario, our approach achieves over 80\% accuracy despite the complete absence of real images for the test indenters. It highlights the strong sim-to-real generalization capability of DOT-Sim.


\begin{table}
\centering
\begin{tabular}{lccc}
\toprule
\textbf{Method} & \textbf{Skin 1} & \textbf{Skin 2} & \textbf{Skin 3} \\
& Acc. \% & Acc. \% & Acc. \% \\
\midrule
DiffTactile & 52.78 & 46.15 & 51.72 \\
Tacto & 38.89 & 46.15 & 44.83 \\
\textbf{DOT-Sim} & \textbf{80.56} & \textbf{92.31} & \textbf{96.55} \\
\bottomrule
\end{tabular}
\caption{Tumor classification results.}
\label{tab:tumor_classification}
\vspace{-2.5em}
\end{table}

\begin{figure*}[ht!]
\centering
\includegraphics[width=0.9\linewidth]{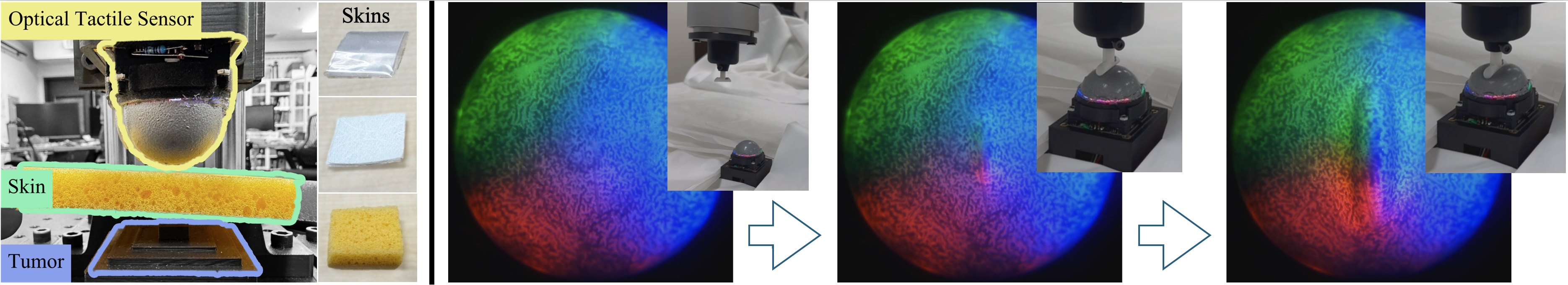}
\caption{Experimental setup for tumor detection (left) and sim-to-real trajectory following (right).}
\label{fig:tumor_and_traj}
\vspace{-2em}
\end{figure*}

\paragraph{Tumor Detection via Tactile Feedback} Given a tactile image resulting from pressing on soft skin, this task is to classify the presence of a tumor. We simulate the presence of a tumor using a convex shape, and its absence using a concave shape, as illustrated on the left side of Figure~\ref{fig:tumor_and_traj}. \revised{Skin deformation is simulated using a position-based dynamics (PBD) solver, for balance of speed and realistic elastic behavior.} A tactile classifier is trained exclusively on synthetic data generated by DOT-Sim and evaluated on real-world images across three levels of skin stiffness, simulated using foams of varying thickness. For evaluation, we collect 36, 13, and 29 real tactile images corresponding to each stiffness level for Skin 1, Skin 2, Skin 3 respectively. Table~\ref{tab:tumor_classification} shows the results (note that for Skin 2, DiffTactile and Tacto have the same 46.15\% accuracy, since they both get 6/13 correct).


\paragraph{Precise Trajectory Following with Sim-to-Real Transfer}

\revised{
To demonstrate how DOT-Sim enables real-world control, we perform trajectory-following via sim-to-real transfer, where an agent mimics simulated demonstrations in a physical environment. Our policy relies solely on synthetic tactile images as input, without requiring force measures.
}

\revised{
We train a ResNet-18 policy via behavior cloning in simulation, which outputs a 6-DoF velocity command from each tactile image, consisting of translational and angular velocities, which is sent to the robot’s Cartesian controller.
}

\revised{
For real-time deployment on the xArm 7, we implement a multi-threaded control system with three threads: (1) image acquisition at 25~Hz, (2) policy inference ($\sim$3.9 ms) followed by synchronization to maintain 25~Hz, and (3) Cartesian control with 100~Hz pose feedback. The low latency ensures synchronization across sensing, inference, and control.
}

Qualitative results are shown in Figure~\ref{fig:tumor_and_traj}. The transferred policy successfully tracks the demonstration, achieving an average action error of $0.896 \pm 0.031$ mm over 10 trials.  



\subsection{Reinforcement Learning}

\revised{Although DOT-Sim primarily targets accurate physical and optical simulation, we also demonstrate its utility for control via reinforcement learning. We consider a box-repositioning task: using only DenseTact tactile images, the agent must rotate the box to a target yaw of $10^{\circ}$ (i.e., flip by $10^{\circ}$), while maintaining stable contact. Success is declared when the yaw error drops below $<\!2^{\circ}$.}

\begin{figure*}[ht]
    \centering
    \vspace{0.7em}
    \begin{subfigure}[t]{0.2\linewidth}
        \centering
        \includegraphics[width=\linewidth]{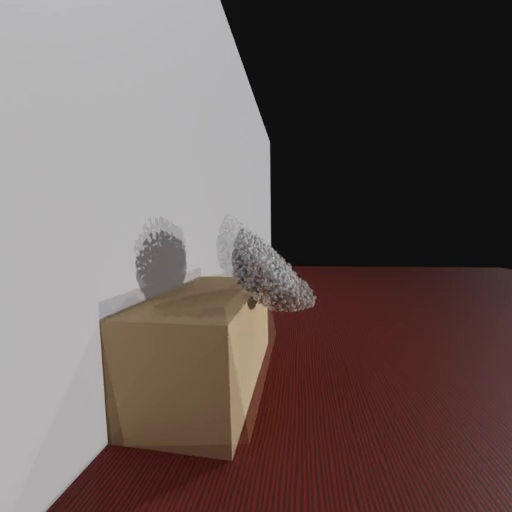}
        \caption{$t_0$}
    \end{subfigure}
    \begin{subfigure}[t]{0.2\linewidth}
        \centering
        \includegraphics[width=\linewidth]{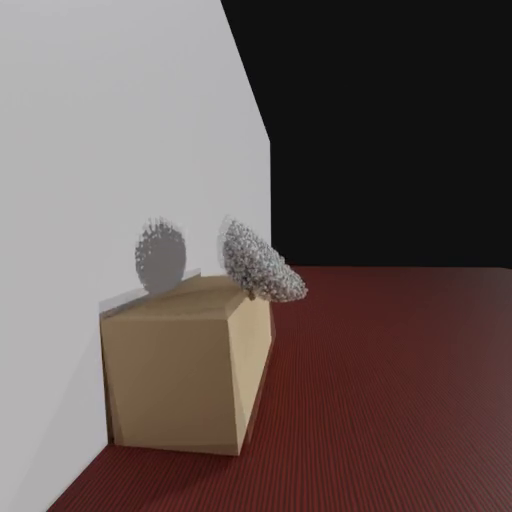}
        \caption{$t_1$}
    \end{subfigure}
    \begin{subfigure}[t]{0.2\linewidth}
        \centering
        \includegraphics[width=\linewidth]{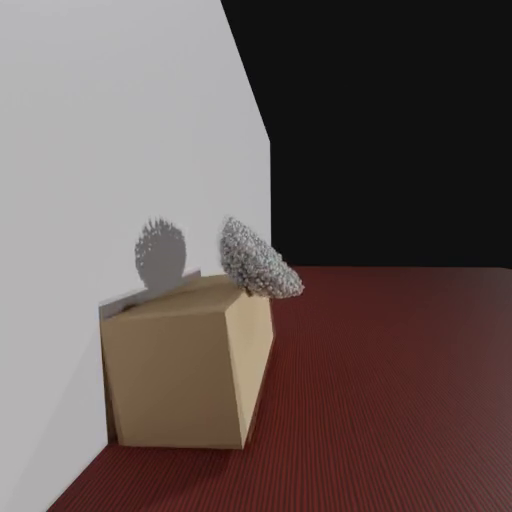}
        \caption{$t_2$}
    \end{subfigure}
    \begin{subfigure}[t]{0.2\linewidth}
        \centering
        \includegraphics[width=\linewidth]{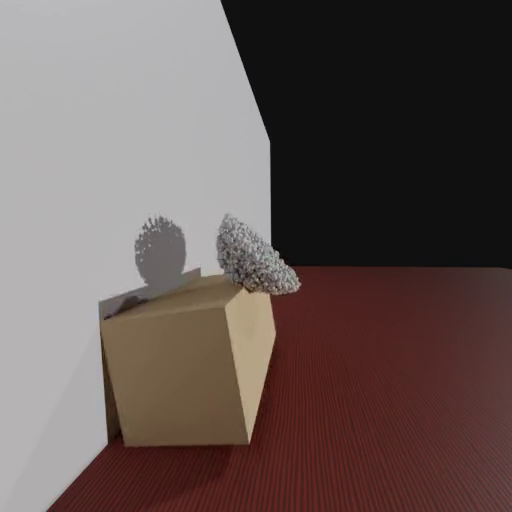}
        \caption{$t_3$}
    \end{subfigure}
    \caption{Policy rollout snapshots (left to right). DenseTact learns to drive the box toward the target yaw. The target position is overlaid semi-transparently.}
    \vspace{-1em}
    \label{fig:rl_seq}
\end{figure*}

\revised{We train a PPO~\cite{schulman_proximal_2017} agent with SKRL~\cite{serrano2023skrl} library in our differentiable DenseTact environment with a discrete \mbox{2-DoF} action space. At every timestep the agent observes a \mbox{3-channel} optical image simulated by our algorithm, flattened and encoded by a ResNet-18; the policy (categorical) and value heads share this backbone. The action space consists of 9 planar velocity commands \{stay, left, right, up, down, up-left, up-right, down-left, down-right\} with a fixed action magnitude. The reward primarily encourages reaching a target yaw of $10^{\circ}$ about the sensor axis, with success defined as an angular error $<\!2^{\circ}$; additional shaping penalizes ineffective no-op actions when far from the goal. Episodes are capped at 1s (24 FPS). During training and evaluation we use the calibrated physical parameters (e.g., $E\!=\!27575$, $\nu\!=\!0.303$) as specified in Section~\ref{sec:physics}. The policy converges quickly (within 15 minutes). Figure~\ref{fig:rl_seq} visualizes a successful rollout, showing the executed over time.}

\section{Conclusion and Limitations} 
We presented DOT-Sim, a differentiable simulation framework for optical tactile sensors that couples deformation modeling via the Material Point Method with a residual-based optical rendering pipeline. DOT-Sim achieves physically and visually consistent simulations using minimal real-world data, and transfers effectively to downstream tasks such as classification and control. Its modular design further allows integration with any MPM-based engine through a lightweight optical rendering plugin.  

Despite these strengths, DOT-Sim has several limitations. First, it struggles to generalize to highly out-of-distribution geometries, especially those with sharp edges or fine surface details, leading to imperfect alignment between simulated and real contacts. Potential remedies include improving MPM efficiency to support denser point sampling, incorporating residual models to refine local deformations, and enlarging the indenter dataset to better capture geometric diversity. Second, the current simulation pipeline is computationally demanding, running at approximately 3~FPS on an NVIDIA A6000 GPU, which restricts its use in real-time control. \revised{While our reported configuration does not achieve high FPS, runtime can be significantly improved by tuning MPM parameters such as voxel resolution, contact density, and substeps per frame, yielding substantially higher frame rates with only marginal increases in error (see Table~\ref{tab:runtime}).}

\begin{table}[h]
\centering
\resizebox{\linewidth}{!}{
\begin{tabular}{ccccc}
    \toprule
    Voxel res. (mm) & Softness & \# substeps & FPS & PSNR (Medium) \\
    \midrule
    \textbf{1.2} & \textbf{15} & \textbf{100} & \textbf{3.6} & \textbf{31.39} \\
    1.2 & 15 & 20 & 17.1 & 30.17 \\
    \midrule
    2.4 & 30 & 100 & 3.8 & 30.98 \\
    2.4 & 30 & 20 & 17.2 & 29.79 \\
    \bottomrule
\end{tabular}}
\caption{Runtime and accuracy trade-offs with different MPM parameters. Contact density is parameterized by \emph{softness}, where particles within $1 / \text{\emph{softness}}$ are considered in contact. The first row corresponds to the parameter settings used in the main experiments.}
\vspace{-2em}
\label{tab:runtime}
\end{table}

\bibliographystyle{IEEEtran}
\bibliography{main}  

@article{wang2022tacto,
  title={Tacto: A fast, flexible, and open-source simulator for high-resolution vision-based tactile sensors},
  author={Wang, Shaoxiong and Lambeta, Mike and Chou, Po-Wei and Calandra, Roberto},
  journal={IEEE Robotics and Automation Letters},
  volume={7},
  number={2},
  pages={3930--3937},
  year={2022},
  publisher={IEEE}
}

@article{serrano2023skrl,
  author  = {Antonio Serrano-Muñoz and Dimitrios Chrysostomou and Simon Bøgh and Nestor Arana-Arexolaleiba},
  title   = {skrl: Modular and Flexible Library for Reinforcement Learning},
  journal = {Journal of Machine Learning Research},
  year    = {2023},
  volume  = {24},
  number  = {254},
  pages   = {1--9},
  url     = {http://jmlr.org/papers/v24/23-0112.html}
}

@manual{Dassault2024Abaqus,
  title        = {Abaqus/Standard and Abaqus/Explicit User’s Manual},
  author       = {{Dassault Syst{\`e}mes Simulia Corp.}},
  organization = {Dassault Syst{\`e}mes SIMULIA},
  address      = {Providence, RI, USA},
  year         = {2024},
  edition      = {2024},
  url          = {https://www.3ds.com/products/simulia/abaqus/}
}

@article{chen2022bidirectional,
  title={Bidirectional sim-to-real transfer for gelsight tactile sensors with cyclegan},
  author={Chen, Weihang and Xu, Yuan and Chen, Zhenyang and Zeng, Peiyu and Dang, Renjun and Chen, Rui and Xu, Jing},
  journal={IEEE Robotics and Automation Letters},
  volume={7},
  number={3},
  pages={6187--6194},
  year={2022},
  publisher={IEEE}
}

@inproceedings{hu2019chainqueen,
  title={Chainqueen: A real-time differentiable physical simulator for soft robotics},
  author={Hu, Yuanming and Liu, Jiancheng and Spielberg, Andrew and Tenenbaum, Joshua B and Freeman, William T and Wu, Jiajun and Rus, Daniela and Matusik, Wojciech},
  booktitle={2019 International conference on robotics and automation (ICRA)},
  pages={6265--6271},
  year={2019},
  organization={IEEE}
}

@article{hu2019difftaichi,
  title={DiffTaichi: Differentiable Programming for Physical Simulation},
  author={Hu, Yuanming and Li, Ziheng and Anderson, Luke and Kaya, Tzu-Mao Li and Ritchie, Daniel and Durand, Frédo and Freeman, William T and Tenenbaum, Joshua B and Matusik, Wojciech},
  journal={International Conference on Learning Representations (ICLR)},
  year={2020}
}

@inproceedings{zhang2024physdreamer,
  title={Physdreamer: Physics-based interaction with 3d objects via video generation},
  author={Zhang, Tianyuan and Yu, Hong-Xing and Wu, Rundi and Feng, Brandon Y and Zheng, Changxi and Snavely, Noah and Wu, Jiajun and Freeman, William T},
  booktitle={European Conference on Computer Vision},
  pages={388--406},
  year={2024},
  organization={Springer}
}

@article{yuan2017gelsight,
  title={Gelsight: High-resolution robot tactile sensors for estimating geometry and force},
  author={Yuan, Wenzhen and Dong, Siyuan and Adelson, Edward H},
  journal={Sensors},
  volume={17},
  number={12},
  pages={2762},
  year={2017},
  publisher={MDPI}
}

@inproceedings{johnson2009retrographic,
  title={Retrographic sensing for the measurement of surface texture and shape},
  author={Johnson, Micah K and Adelson, Edward H},
  booktitle={2009 IEEE Conference on Computer Vision and Pattern Recognition},
  pages={1070--1077},
  year={2009},
  organization={IEEE}
}

@article{lambeta2020digit,
  title={Digit: A novel design for a low-cost compact high-resolution tactile sensor with application to in-hand manipulation},
  author={Lambeta, Mike and Chou, Po-Wei and Tian, Stephen and Yang, Brian and Maloon, Benjamin and Most, Victoria Rose and Stroud, Dave and Santos, Raymond and Byagowi, Ahmad and Kammerer, Gregg and others},
  journal={IEEE Robotics and Automation Letters},
  volume={5},
  number={3},
  pages={3838--3845},
  year={2020},
  publisher={IEEE}
}

@inproceedings{do2022densetact,
  title={Densetact: Optical tactile sensor for dense shape reconstruction},
  author={Do, Won Kyung and Kennedy, Monroe},
  booktitle={2022 International Conference on Robotics and Automation (ICRA)},
  pages={6188--6194},
  year={2022},
  organization={IEEE}
}

@inproceedings{do2023densetact,
  title={Densetact 2.0: Optical tactile sensor for shape and force reconstruction},
  author={Do, Won Kyung and Jurewicz, Bianca and Kennedy, Monroe},
  booktitle={2023 IEEE International Conference on Robotics and Automation (ICRA)},
  pages={12549--12555},
  year={2023},
  organization={IEEE}
}

@inproceedings{
    xu2022efficient,
    title={Efficient Tactile Simulation with Differentiability for Robotic Manipulation},
    author={Jie Xu and Sangwoon Kim and Tao Chen and Alberto Rodriguez Garcia and Pulkit Agrawal and Wojciech Matusik and Shinjiro Sueda},
    booktitle={6th Annual Conference on Robot Learning},
    year={2022},
    url={https://openreview.net/forum?id=6BIffCl6gsM}
}

@inproceedings{gao2016deep,
  title={Deep learning for tactile understanding from visual and haptic data},
  author={Gao, Yang and Hendricks, Lisa Anne and Kuchenbecker, Katherine J and Darrell, Trevor},
  pages={536--543},
  year={2016},
  organization={IEEE}
}

@inproceedings{xie2020grnet,
  title={Grnet: Gridding residual network for dense point cloud completion},
  author={Xie, Haozhe and Yao, Hongxun and Zhou, Shangchen and Mao, Jiageng and Zhang, Shengping and Sun, Wenxiu},
  booktitle={European conference on computer vision},
  pages={365--381},
  year={2020},
  organization={Springer}
}

@inproceedings{yuan2018pcn,
  title={Pcn: Point completion network},
  author={Yuan, Wentao and Khot, Tejas and Held, David and Mertz, Christoph and Hebert, Martial},
  booktitle={2018 international conference on 3D vision (3DV)},
  pages={728--737},
  year={2018},
  organization={IEEE}
}

@article{zhong2024springgaus,
    title     = {Reconstruction and Simulation of Elastic Objects with Spring-Mass 3D Gaussians},
    author    = {Zhong, Licheng and Yu, Hong-Xing and Wu, Jiajun and Li, Yunzhu},
    journal   = {European Conference on Computer Vision (ECCV)},
    year      = {2024}
}

@inproceedings{agarwal2021simulation,
  title={Simulation of Vision-based Tactile Sensors using Physics based Rendering},
  author={Agarwal, Arpit and Man, Timothy and Yuan, Wenzhen},
  booktitle={Proceedings of the International Conference on Robotics and Automation (ICRA)},
  year={2021},
  doi={10.1109/ICRA48506.2021.9561122},
  url={https://scispace.com/papers/simulation-of-vision-based-tactile-sensors-using-physics-4lcow7l4d5}
}

@article{si2024difftactile,
  title={DIFFTACTILE: A Physics-based Differentiable Tactile Simulator for Contact-rich Robotic Manipulation},
  author={Si, Zilin and Zhang, Gu and Ben, Qingwei and Romero, Branden and Zhou, Xian and Liu, Chao and Gan, Chuang},
  journal={arXiv preprint arXiv:2403.08716},
  year={2024},
  doi={10.48550/arxiv.2403.08716},
  url={https://scispace.com/papers/difftactile-a-physics-based-differentiable-tactile-simulator-1q4pbz4xgc}
}

@article{swann2024touchgs,
  author    = {Aiden Swann and Matthew Strong and Won Kyung Do and Gadiel Sznaier Camps and Mac Schwager and Monroe Kennedy III},
  title     = {Touch-GS: Visual-Tactile Supervised 3D Gaussian Splatting},
  journal   = {arXiv},
  year      = {2024},
  booktitle={2024 IEEE/RSJ International Conference on Intelligent Robots and Systems (IROS)}, 
  pages={10511-10518},
  doi={10.1109/IROS58592.2024.10802412}  
}

@inproceedings{qi2023general,
  title={General in-hand object rotation with vision and touch},
  author={Qi, Haozhi and Yi, Brent and Suresh, Sudharshan and Lambeta, Mike and Ma, Yi and Calandra, Roberto and Malik, Jitendra},
  booktitle={Conference on Robot Learning},
  pages={2549--2564},
  year={2023},
  organization={PMLR}
}

@misc{zhao2024fotsfastopticaltactile,
      title={FOTS: A Fast Optical Tactile Simulator for Sim2Real Learning of Tactile-motor Robot Manipulation Skills}, 
      author={Yongqiang Zhao and Kun Qian and Boyi Duan and Shan Luo},
      year={2024},
      eprint={2404.19217},
      archivePrefix={arXiv},
      primaryClass={cs.RO},
      url={https://arxiv.org/abs/2404.19217}, 
}

@article{si2021taxim,
  title={Taxim: An Example-Based Simulation Model for GelSight Tactile Sensors},
  author={Yuan, Wenzhen and Si, Zilin},
  journal={IEEE Robotics and Automation Letters},
  volume={7},
  number={2},
  pages={3475--3482},
  year={2022},
  doi={10.1109/LRA.2022.3142412},
  url={https://scispace.com/papers/taxim-an-example-based-simulation-model-for-gelsight-tactile-f9075kf7}
}

@incollection{jiang2016material,
  title={The material point method for simulating continuum materials},
  author={Jiang, Chenfanfu and Schroeder, Craig and Teran, Joseph and Stomakhin, Alexey and Selle, Andrew},
  booktitle={Acm siggraph 2016 courses},
  pages={1--52},
  year={2016}
}

@article{chen2017rethinking,
  title={Rethinking atrous convolution for semantic image segmentation},
  author={Chen, Liang-Chieh and Papandreou, George and Schroff, Florian and Adam, Hartwig},
  journal={arXiv preprint arXiv:1706.05587},
  year={2017}
}

@article{schulman_proximal_2017,
    title = {Proximal {Policy} {Optimization} {Algorithms}},
    url = {http://arxiv.org/abs/1707.06347},
    author = {Schulman, John and Wolski, Filip and Dhariwal, Prafulla and Radford, Alec and Klimov, Oleg},
    year = {2017},
    note = {arXiv: 1707.06347},
    pages = {1--12},
}

@book{ogden1997non,
  title={Non-linear elastic deformations},
  author={Ogden, Raymond W},
  year={1997},
  publisher={Courier Corporation}
}

@article{zhao2024ifem2,
  title={iFEM2. 0: Dense 3D Contact Force Field Reconstruction and Assessment for Vision-Based Tactile Sensors},
  author={Zhao, Can and Liu, Jin and Ma, Daolin},
  journal={IEEE Transactions on Robotics},
  year={2024},
  publisher={IEEE}
}

@article{do2025tensortouch,
  title={TensorTouch: Calibration of Tactile Sensors for High Resolution Stress Tensor and Deformation for Dexterous Manipulation},
  author={Do, Won Kyung and Strong, Matthew and Swann, Aiden and Lei, Boshu and Kennedy III, Monroe},
  journal={arXiv preprint arXiv:2506.08291},
  year={2025}
}


\section*{Appendix}

\subsection{Optical Rendering Network Architecture}

Our optical rendering network follows a standard encoder-decoder design based on the DeepLabV3-ResNet50 architecture~\cite{chen2017rethinking}, implemented using the PyTorch torchvision library. The input is a 4-channel image composed of rendered surface normals and depth maps (3 for normal, 1 for depth), both normalized to $[0,1]$. The output is a three-channel RGB image, trained to match ground-truth camera images.

We append a fully connected layer with 3 channels, and train the network using a per-pixel $\ell_2$ loss between predicted and target RGB values. No perceptual or adversarial loss is used. The model is optimized with Adam using a learning rate of $3\times10^{-4}$, weight decay of $1\times10^{-4}$, and a batch size of 8 for 100 epochs. Input images are resized to $640 \times 480$.

We found that the $\ell_2$ loss alone yields sharp reconstructions when using accurate geometry and surface rendering, without requiring additional regularization.
\subsection{Evaluation Metrics}

\paragraph{Physical Accuracy Metrics.} 
To evaluate the accuracy of predicted deformations in terms of 3D geometry, we adopt four standard point cloud comparison metrics, following~\cite{yuan2018pcn, xie2020grnet}. All metrics are computed on point clouds uniformly sampled with 2,048 points from both the predicted and ground-truth surfaces.

\begin{itemize}
    \item \textbf{L2 Chamfer Distance (CD)}: Measures the average distance from each point in one point cloud to its nearest neighbor in the other. It captures overall geometric similarity and is symmetric by averaging both directions.
    \item \textbf{Significant L2 Chamfer Distance}: A variant of Chamfer Distance that focuses on the top 1\% largest nearest-neighbor distances for each point, emphasizing outliers and surface discrepancies.
    \item \textbf{Earth Mover's Distance (EMD)}: Computes the optimal bijective assignment between points in the two clouds that minimizes total transport cost. EMD is more sensitive to global shape structure.
    \item \textbf{F-Score @ 1mm}: Computes the harmonic mean of precision and recall under a \textit{1mm} distance threshold. A predicted point is considered a true positive if it lies within \textit{1mm} of any ground-truth point, and vice versa.
\end{itemize}

These metrics jointly evaluate both average-case performance (CD, EMD) and worst-case or perceptual differences (Sig. CD, F-Score), offering a balanced picture of physical prediction fidelity.

\paragraph{Optical Rendering Accuracy Metrics.}
To quantify the accuracy of rendered tactile images, we compute:

\begin{itemize}
    \item \textbf{Mean L2 Norm} ($\downarrow$): The average pixel-wise L2 norm difference between predicted and ground-truth RGB values, normalized to $[0,1]$.
    \item \textbf{Significant Pixel L2 Norm} ($\downarrow$): Similar to the physical Sig. CD, we compute the average L2 norm over the top 1\% worst-predicted pixels to highlight regions with high rendering error.
    \item \textbf{Peak Signal-to-Noise Ratio (PSNR)} ($\uparrow$): A standard perceptual metric that expresses the ratio between the maximum possible pixel value and the reconstruction error. Let $\hat{I}, I \in [0,1]^{H \times W \times 3}$ denote the predicted and ground-truth RGB images. First compute the mean squared error (MSE) as $\|\hat{I}-I\|_2^2$, and then PSNR is given by:
    \[
    \text{PSNR} = 10 \cdot \log_{10} \left( \frac{1}{\text{MSE}} \right).
    \]
\end{itemize}


\end{document}